# A Systematic Review on Prompt Engineering in Large Language Models for K-12 STEM Education

I-SHENG CHEN*, Carnegie Mellon University, USA
DANYANG WANG*, University of Sheffield, UK
LUYI XU*, Harvard University, USA
CHEN CAO, University of Sheffield, UK
XIAO FANG, Massachusetts Institute of Technology, USA
JIONGHAO LIN†, Carnegie Mellon University, USA

Large language models (LLMs) have the potential to enhance K-12 STEM education by improving both teaching and learning processes. While previous studies have shown promising results, there is still a lack of comprehensive understanding regarding how LLMs are effectively applied, specifically through prompt engineering—the process of designing prompts to generate desired outputs. To address this gap, our study investigates empirical research published between 2021 and 2024 that explores the use of LLMs combined with prompt engineering in K-12 STEM education. Following the PRISMA protocol, we screened 2,654 papers and selected 30 studies for analysis. Our review identifies the prompting strategies employed, the types of LLMs used, methods of evaluating effectiveness, and limitations in prior work. Results indicate that while simple and zero-shot prompting are commonly used, more advanced techniques like few-shot and chain-of-thought prompting have demonstrated positive outcomes for various educational tasks. GPT-series models are predominantly used, but smaller and fine-tuned models (e.g., Blender 7B) paired with effective prompt engineering outperform prompting larger models (e.g., GPT-3) in specific contexts. Evaluation methods vary significantly, with limited empirical validation in real-world settings.

## 1 INTRODUCTION

K-12 education refers to the publicly supported school grades prior to college in the United States and several other countries. The term "K-12" stands for "Kindergarten through 12th grade" and represents the full range of primary and secondary education. Within this system, a strong emphasis has been placed on STEM (**S**cience, **T**echnology, **E**ngineering, and **M**athematics) education as a means to prepare students for a technology-driven future. STEM education at the K-12 level focuses on building foundational knowledge in scientific inquiry, technological literacy, engineering principles, and mathematical reasoning [10, 29, 64]. The K-12 STEM education emphasizes interdisciplinary learning, where students apply concepts from multiple domains to solve real-world challenges, such as integrating mathematics with science to tackle engineering problems [29]. The importance of K-12 STEM education lies in its ability to prepare students for a rapidly evolving, technology-driven world by fostering critical thinking, creativity, and problem-solving skills from an early age [10]. Students who engage in well-structured STEM curricula are more likely to pursue further education and careers in high-demand fields like information technology and engineering which are essential for technological innovation. Additionally, K-12 STEM education equips students with competencies such as analytical thinking, which prepare them for a wide range of career paths while enabling them to tackle complex problems [64]. Recognizing the importance of STEM education at the K-12 level, it is essential to deliver K-12 STEM education at scale to ensure equitable access to individual students. Achieving this scalability requires the integration of advanced technologies that can enhance learning experiences, increase accessibility, and adapt to the diverse needs of students across different educational contexts.

---

*These authors contributed equally to this work.
†Corresponding author.

Many previous works have focused on developing computer systems and artificial intelligence technologies to support K-12 STEM education [59]. These technological advancements have demonstrated significant potential and efficacy in enhancing student learning. For example, Intelligent Tutoring Systems (ITS) are computer-based systems that provide personalized instruction and feedback to students based on their individual learning needs [12, 61]. In K-12 STEM education, ITS has been widely adopted across various K-12 STEM subjects, such as math [22, 33], physics [12, 60], and programming [41]. Despite these promising developments, there remains a need for continued technological improvement in K-12 STEM education to fully meet the diverse needs of learners and ensure equitable access to high-quality STEM education. Recent advancements in artificial intelligence (AI), like Large Language Models (LLMs), offer the potential to further enhance how K-12 STEM education is delivered and supported, enabling personalized and accessible learning experiences at scale [32, 87].

LLMs are AI models trained on vast amounts of text data to generate human-like text, comprehend context, and perform a wide range of tasks across various domains [31]. The LLM models, such as GPT-4 [1] and Claude 3.5 [4], are capable of generating coherent and contextually relevant responses, making them valuable tools in STEM education [8, 74]. To interact with LLMs, users need to provide a segment of guiding text, known as a "prompt." A prompt serves as the input that instructs the model on how to respond, shaping the model's output based on the specific task at hand [9]. Effective prompts are crucial for eliciting accurate, relevant, and contextually appropriate responses from the model [4, 34]. Achieving an effective prompt often requires a process known as prompt engineering, which involves carefully designing, refining, and iterating prompts to optimize the LLM's performance on specific tasks [45, 77]. Prompt engineering entails understanding how the model interprets the input, testing different phrasing, adjusting for specificity, and even incorporating additional steps, such as reasoning or formatting constraints, to improve results [45, 77]. Despite the growing interest in LLMs in education [11, 40, 78], prompt engineering is still in its early stages. There is limited understanding of how LLMs interpret prompts and the underlying mechanisms that affect task performance, making it a developing field in need of further exploration. Recognizing this gap, our study aims to systematically examine how existing research in K-12 STEM education has utilized prompt engineering, the methods they employed to evaluate these prompts, and the challenges or limitations encountered in the field. Additionally, we seek to identify key insights that can inspire future research directions in education and learning analytics.

## 2 BACKGROUND
### 2.1 Large Language Models in K12 STEM Education

Large Language Models (LLMs) are advanced artificial intelligence (AI) models designed to comprehend and generate human-like language by being trained on vast datasets of text [31]. These models are termed "large" due to their extensive number of parameters—often reaching billions—which enable them to capture and replicate complex language patterns with high fidelity [55]. As a result, LLMs have shown potential across many educational tasks [32], such as personalized tutoring [72] and automatic grading [20]. The integration of LLMs into K-12 STEM education holds great promise for improving how students engage with challenging subjects such as mathematics, science, and technology [14, 35]. One of the key benefits of LLMs in this context is their ability to provide real-time feedback. Unlike traditional classroom settings, where feedback from teachers may be delayed, LLMs can offer immediate clarification and guidance, helping students to better understand complex material as they encounter difficulties. This instant feedback loop is particularly valuable in subjects like mathematics and physics, where students often need immediate assistance to grasp abstract concepts and solve problems efficiently [48, 72]. Despite the promise LLMs hold for improving K-12 STEM education, there are also limitations and challenges to consider. For instance, while LLMs are capable of generating coherent and accurate responses, they are not immune to errors, and their outputs may sometimes lack depth or accuracy, especially when it comes to highly specialized or domain-specific knowledge [25]. Therefore, the process of designing prompts to guide LLMs for content generation is important, and human oversight remains critical to ensure the effective use of LLMs in educational contexts.

## 2.2 Prompt Engineering

Prompting refers to the specific input or instructions provided to LLMs to guide their output generation and shape their responses to various tasks and queries [9, 47, 65, 67]. Effective prompt engineering is crucial, as it directly influences the relevance and accuracy of the LLM's outputs [47, 65, 67]. Prompt engineering involves the systematic crafting and refining of prompts to optimize the model's performance and generate desired outputs [47, 65, 67]. According to various studies [65], prompting strategies can be categorized based on their purpose, such as prompting LLMs without extensive model training (e.g., Zero-shot [63] and Few-shot [9]), prompting LLMs to demonstrate reasoning and logic (e.g., Chain-of-Thoughts [80]), and prompting LLMs to mitigate hallucinations (e.g., Retrieval-Augmented Generation [37]). As the goal of this study is to examine the role of prompting strategies in STEM education, we primarily focus on the prompting strategies applied to previous works in K-12 STEM education. **Simple Prompting (Direct Instruction Prompting)**: Users explicitly instruct LLMs to generate desired outputs without additional context or examples [65, 67]. For example, prompts like "*Solve this algebraic equation*" or "*Explain Newton's third law*" direct the model to respond directly to the query. **Role-assigned Prompting (Persona-Based Prompting)**: Instructing the LLM to adopt a specific role or persona, such as a teacher, student, or domain expert, to generate responses aligned with that role [67, 79, 88]. For example, "*Act as a high school physics teacher*" prompts the model to simulate how a teacher might explain complex topics, guide students, or provide feedback. **Zero-Shot Prompting**: The model is given a task and expected to generate a response based on its general knowledge, without prior examples or specific training for that task [63]. **Few-Shot Prompting**: Providing the model with a small number of examples to guide its responses, enhancing its understanding of the task and leading to more accurate outputs [9]. For instance, include solved examples of a problem before presenting a new, unsolved one. **Chain of Thought (CoT) Prompting**: Encouraging the model to reason through problems step-by-step, improving logical coherence and accuracy [80]. This can be applied in both zero-shot and few-shot contexts and is valuable for tasks requiring complex reasoning. **Retrieval-Augmented Generation (RAG)**: LLMs can have limitations, such as hallucinations, contradictory responses, incorrect reasoning, or outdated training data [6, 7, 15, 17, 23]. Combining LLM prompting with external information retrieval mechanisms allows the model to access external knowledge bases to retrieve and output relevant and up-to-date information [37] and further improve accuracy [39]. **Output Formatting**: Designing prompts that guide LLMs to generate responses in a structured, organized format, facilitating programmatic processing or quantitative analysis [67, 83].

## 2.3 Existing works on systematic literature review of LLM in education

Prior systematic literature reviews have primarily focused on potentials of LLM applications [11, 40, 78, 85]. For example, [85] summarized the pracitcial applications of LLMs to differnt types of educational tasks including grading, teaching support, and feedback. However, the study [85] mainly focus on the applications of LLMs prior to the advanced LLM models (e.g., GPT-4 and Claude) and did not in detail how the use of prompting strategies in delivering desired generated output. Similar to other systematic literature reviews on the use of LLMs in education [5, 19, 58, 85], rare of them have specifically focused on the usage and impact of **prompt engineering strategies** in K-12 STEM educarion. To this end, our study seeks to address this gap by systematically reviewing the role of prompt engineering in K-12 STEM education.

While many previous studies [49, 81, 86] prompted LLMs to operate certain tasks related to K-12 STEM education, the question about how prompt engineering can best be leveraged to enhance learning outcomes still remains largely unknown as the the applications of LLM in K-12 STEM education remains in its early stages. Recognizing the significance of **prompt engineering** for educational purposes, our study aims to conduct a **systematic literature review (SLR)** that delves into the comprehensive effectiveness, strengths, and limitations of different prompt strategies in K-12 education. This includes evaluating their impact on learning outcomes, how they enhance interactions between students, instructors, and LLM-based systems, and the practical challenges faced when implementing these strategies in educational settings. By thoroughly examining existing research and current trends, our study aims to provide a detailed understanding of how prompt engineering can be further refined and applied to optimize its potential in K-12 STEM education. Thus, our research seeks to address the following **R**esearch **Q**uestions (**RQ**):

**RQ 1:** What methods are used for implementing LLMs in K-12 STEM education?
    **RQ 1.1:** What specific prompting strategies have been used with LLMs in previous studies?
    **RQ 1.2:** What types of LLMs are used in previous studies?
    **RQ 1.3:** How did previous studies build LLM-based systems to facilitate K-12 STEM education?
**RQ 2:** How is the effectiveness of LLMs in K-12 STEM education evaluated?
    **RQ 2.1:** What methods are used to evaluate the performance of LLMs with prompting strategies?
    **RQ 2.2:** What metrics are considered for measuring LLM performance with prompting strategies?
    **RQ 2.3:** How have previous studies validated LLM prompting strategies with real users?
**RQ 3:** What are the limitations of previous research on the use of LLMs in K-12 education?

## 3 METHODS
### 3.1 Review Procedures

We followed the PRISMA [56] protocol to conduct our current systematic review on the use of prompt engineering with large language models in STEM education. A comprehensive search was conducted across several reputable databases, including Scopus, ACM Digital Library, IEEE Xplore, Web of Science (WoS), and Education Resources Information Center (ERIC) from ProQuest to ensure the inclusion of high-quality peer-reviewed publications. In addition, we performed supplementary searches through Arxiv and SSRN from Socpus (preprint) to capture publications that may not yet published. Our initial search query used was: `TITLE-ABS-KEY ( ( chatgpt OR large AND language AND model ) AND prompt AND education AND ( science OR technology OR engineering OR math OR steam ) ) AND PUBYEAR > 2021 AND PUBYEAR < 2025`

Given the significant growth and development of prompt engineering and GPT [51] after 2021, we review papers from 2021 to January 25, 2024. To expand the range of potential sources and consider papers that have not yet been accepted or published, we included both peer-reviewed papers and preprints found through Scopus preprint searches. The initial search yielded 3,095 publications. After removing 441 duplicates, 2,654 papers remained for title and abstract screening. Our reviewing process is shown at Figure 1.

Notably, our study only considers papers written in English. A set of inclusion and exclusion criteria were used to select the relevant studies from the initial inquiry. First, studies that do not mention prompt engineering or use LLMs were excluded. Additionally, research unrelated to Generative AI was excluded, such as only using BERT for regression or classification. Third, papers that do not focus on STEM education were also excluded.

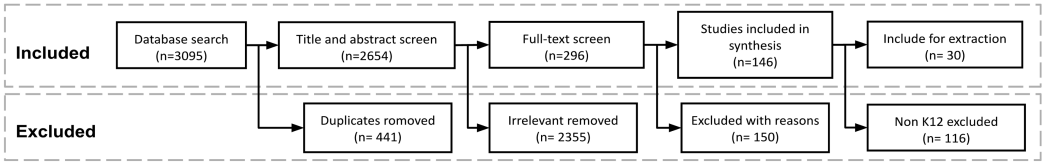

Fig. 1. Systematic scoping review process following the PRISMA protocol.

Six reviewers with experience in education and computer science reviewed the titles and abstracts of the studies based on inclusion and exclusion criteria. After screening, 296 papers were included for the next step with a with an absolute percentage agreement score of 76.8%, indicating a moderate to good agreement between the reviewers[68] . The mild moderate agreement stems from differing views on the scope of STEM education, such as whether fields like statistics, medicine, and finance should be included in STEM. Ultimately, we refined the definition of STEM for this paper based on the inclusion criteria, ensuring clarity on the fields covered. Moreover, this issue was resolved later once we narrowed our scope to K-12 education. Following the full-text review, 146 papers were initially included for synthesis. However, due to the large number of papers remaining, we decided to focus exclusively on K12 education. As a result, 116 papers were excluded for non-K12 education. This left a final total of 30 studies, all of which met the inclusion criteria.

### 3.2 Data Anylysis

We conducted an in-depth qualitative analysis of the selected studies from three aspects to address the RQs described in Section 2.3. For **RQ1**, we gather information about the models and the prompt engineering strategies used in the study while also checking if the prompts are provided in the paper. Specifically, we explore the role of prompting LLMs. For **RQ2**, we check how these papers evaluate the effectiveness of their prompts with LLMs. As for **RQ3**, we focus on identifying the limitations discussed in these papers.

## 4 RESULTS

Table 1 provides an overview of references that apply LLMs to K-12 STEM education across various subjects, categorized by educational levels. The Table 1 shows that math has the most extensive research focus, particularly at the elementary and high school levels, with numerous studies applying LLMs to support learning tasks in this subject. Subjects like physics, chemistry, biology, and computer sciences are also represented, though to a lesser extent, with more studies occurring at the high school level as students engage with more specialized content. Additionally, interdisciplinary subjects, including psychology, environmental science, and general sciences, are present but show fewer applications of LLMs across all educational levels. Overall, the table highlights that while the application of LLMs has primarily focused on the math domain, there is growing interest in extending these models to support a wider range of STEM subjects.

### 4.1 Commonly used prompting strategies in K-12 STEM education contexts (RQ 1.1)

In our study, we identified various prompting strategies employed in previous studies under the K-12 STEM education context, as shown in Table 2. These strategies include simple prompting, role-assigned prompting, zero-shot prompting, few-shot prompting, chain of thought prompting, retrieval-augmented generation, and output formatting, which were introduced in Section 2.2. These strategies are often compatible and can be combined to use together.

*4.1.1 Simple Prompting.* Simple prompting, or direct instruction prompting, was used in 12 of the reviewed articles. This straightforward approach involves explicitly asking LLMs to generate desired

Table 1. Overview of research studies across different subject areas and educational levels.

| Educational Levels | Math | Physics | Chemistry | Biology | Computer Sciences | Other* |
|---|---|---|---|---|---|---|
| **Elementary School** (K - K5, 6 - 10 years old) | [16, 43, 53, 54, 57, 73, 76] | [16] | [16] | [16] | [2, 71, 73] | [2, 16, 42] |
| **Middle School** (K6 - K8, 11 - 14 years old) | [16, 26, 43, 53, 54, 69, 84] | [13, 16, 18, 27] | [16, 26] | [16] | [2, 70] | [2, 13, 16, 36, 42] |
| **High School** (K9 - K12, 14 - 18 years old) | [16, 43, 53, 54, 69, 82] | [3, 7, 16, 21, 27, 50] | [16] | [16, 50] | [2, 50] | [2, 16, 42] |

*Note:* Other subjects involved are Psychology[13, 16]; General Sciences [2]; Natural Science and Social Science [42]; Physical Sciences [36]; Environmental Science [13].

Table 2. Types of prompts used in LLMs for different educational tasks in K-12 STEM education research.

| Educational Tasks | Simple Prompt | Role-assigned Prompt | Zero Shot | Few Shot | CoT | RAG | Output Formatting |
|---|---|---|---|---|---|---|---|
| Problem Solving | [2, 15, 18, 21, 23, 38, 43, 53, 62, 82] | [2, 53] | [2, 15, 18, 21, 23, 38, 42–44, 52, 53, 70, 82] | [3, 18, 21, 23, 27, 42, 52, 62] | [3, 18, 21, 27, 38, 42, 43] | | [15, 18, 38, 42, 43, 53, 62, 70] |
| Assessment and Grading | [50, 53] | [53] | [36, 50, 53] | [36] | [36] | | [36, 53] |
| Problem Creation | [13, 15, 16, 18, 26] | | [15, 18, 54, 57, 84] | [3, 13, 16, 18, 26, 54, 57, 69, 84] | [3, 18, 54] | [16] | [15, 16, 18, 84] |
| Knowledge Representation | [18] | | [18, 42] | [18, 42] | [18, 42] | | [18, 42] |
| Analyzing and Labeling | [53] | [53] | [28, 42, 53, 76] | [28, 42, 76] | [28, 42] | | [42, 53] |
| Programming Support | | | [70] | | | | [70] |
| Feedback Generation | [2, 16, 43, 50, 53, 62, 73, 82] | [2, 53, 73] | [2, 43, 50, 53, 73, 82] | [16, 62] | [43] | [16] | [16, 43, 53, 62] |
| Lesson Planning | [15, 26, 71] | [71] | [15] | [26] | | | [15] |
| Chatbot | [2, 7, 16, 23, 43, 73, 82] | [2, 7, 73] | [2, 7, 23, 43, 73, 82] | [16, 23] | [43] | [16] | [7, 16, 43] |

outputs without additional context or examples. For instance, [13] used GPT-3.5 for similar problem creation by prompting the model to "*paraphrase this sentence.*" Similarly, [26] asked ChatGPT to "*design a lesson plan for an eighth-grade class focusing on square roots and cube roots.*"

Most studies employing simple prompting explored how LLMs could support educators by generating educational materials, such as lesson planning [26, 71], or by addressing subject-specific inquiries in STEM fields [7, 15]. Our result indicated that while simple prompting is intuitive and accessible, it may not always produce outputs that fully meet user expectations or standards of accuracy. To be specific, several studies noted that LLM outputs generated via simple prompting could be too narrow and cannot consider different aspects [53] or require significant post-generation review, editing, and iterating by human expert to reach acceptable quality [7, 15, 23, 26, 71].

*4.1.2 Role-assigned Prompting.* Role-assigned prompting, or persona-based prompting, was applied in 7 of the reviewed articles. This strategy involves instructing LLMs to assume specific roles to generate responses aligned with those roles. Common roles assigned were *teachers* and *students*. For example, [53] prompted the LLM with "*Suppose you are a middle school math teacher,*" resulting in the model providing guidance and encouragement akin to a teacher. Studies used teacher roles for lesson planning [71] and student guidance [2, 16, 53, 73], while student roles were used to simulate student interactions for teacher training or dataset construction [7, 43].

Role-assigned prompting allows LLMs to generate more contextually relevant and personalized responses but requires a careful prompt design to avoid unintended behaviors. Take "*Suppose you are a middle school math teacher.*"[53] as an example; the research found that even when the prompt also instructs LLMs to share answers, LLMs with role-assigned prompts tend to guide students and avoid giving answers away. This potential issue may be alleviated by multiple descriptive elements to facilitate LLMs to generate more personalized and contextually relevant responses [2].

*4.1.3 Zero-Shot Prompting.* Zero-shot prompting was utilized in 18 of the reviewed articles, primarily in solving math and physics problems. The difference between the Zero-shot and simple prompting techniques is that the zero-shot method asks LLMs to finish a task, while the simple prompt's output is more open-ended. To be specific, the model is tasked with generating solutions based on its pre-trained knowledge without prior examples. For instance, [21, 27, 38, 44] evaluated the efficacy of zero-shot prompting in STEM subjects. Studies found that providing more detailed task descriptions or additional information could enhance the model's performance [28, 36, 38, 53, 70].

*4.1.4 Few-Shot Prompting.* Few-shot prompting was employed in 13 articles. In this article, we integrated the one-shot prompting in this category. By providing the model with a few examples, researchers observed improved performance compared to zero-shot prompting [18, 27, 28, 36, 54, 57, 62, 76]. For instance, [28] found that few-shot prompting led to more accurate and contextually appropriate outputs by helping the model better understand task nuances. Few-shot prompting also helped stabilize output style, producing more standardized responses [36]. Nevertheless, [53] found that smaller models like LLAMA-7B perform worse with five-shot prompting compared to zero-shot, likely due to the model size being too small to handle long prompts, since this issue isn't seen in larger models like LLAMA 70B or GPT-3.5.

*4.1.5 Chain of Thought Prompting.* Chain of Thought (CoT) prompting was used in 8 articles to encourage LLMs to generate intermediate reasoning steps rather than direct answers. This strategy is particularly effective in STEM education, where step-by-step reasoning and comprehensive explanations are essential. Studies applied CoT prompting for content analysis [27, 28, 36], problem-solving [18, 21, 36], and content generation [43].

CoT strategies can be integrated within training data or used in few-shot examples to guide models implicitly towards CoT reasoning (i.e., few-shot CoT) [18]. As training datasets expand, advanced models like GPT-4 exhibit CoT-like reasoning without explicit CoT prompts. For instance, [21] observed GPT-4 naturally breaking down complex physics problems into smaller steps, structuring solutions without explicit CoT instructions. CoT also works effectively across multiple rounds of interaction. [27] improved LLM's performance by having the LLM review and correct its prior output. While [54] applied an iterative CoT approach that asked LLMs to generate Python code solution for the problem first, then ask it to output the answer and explanation. CoT's effectiveness increases when combined with other prompting strategies. For example, [36] found that while CoT alone led to limited improvements in automated grading tasks, combining CoT with more zero-shot task descriptions (e.g., scoring rubrics) significantly enhanced content analysis and output accuracy.

*4.1.6 Retrieval-Augmented Generation (RAG).* Only one study [16] employed RAG to enhance LLM performance in STEM education. The study [16] demonstrated that RAG improved the quality and depth of responses in subjects like math and literature by providing up-to-date or specialized knowledge not included in the LLM's original training data.

*4.1.7 Output Formatting.* Formatted outputs were utilized in 10 articles to guide LLMs in generating responses in structured formats, facilitating further programmatic processing or analysis. For example, some studies instructed the LLM to output data in JSON format for easy parsing [3, 27, 42], while others directed the model to produce code outputs or explicitly label specific types of information [28, 36, 53].

## 4.2 Types of LLMs used with these prompting strategies (RQ 1.2)

The majority of reviewed articles (n = 28) employed models from the GPT series, including GPT-4, GPT-3.5, GPT-3, and GPT-2 [9, 55, 63]. However, a smaller subset of articles (n = 4) utilized alternative models. [16] fine-tuned the LLaMA 13B model [75] to create the EduChat model for question-answering tasks in the Chinese K-12 evaluation suite [30]. Although EduChat outperformed other LLaMA 13B models, it was less effective than GPT-4 and other larger LLMs. [52] used LLaMA-2-7B, LLaMA-2-70B, and BLOOMZ-7.1B-MT alongside GPT series models to assess performance on multiple-choice math questions, concluding that GPT-4 demonstrated the highest performance. [76] compared BLOOM and YouChat against ChatGPT, GPT-3, and traditional machine learning methods in evaluating student responses to open-ended math questions. They found that while GPT-3 outperformed BLOOM and YouChat, all tested LLMs performed worse than traditional machine learning methods.

Interestingly, some studies revealed that smaller, fine-tuned models could outperform larger GPT-series models in specific tasks. [73] showed that a fine-tuned smaller model, Blender 9B, outperformed GPT-3 (175B parameters) in simulating the role of a teacher. Similarly, [16] found that smaller fine-tuned models exhibited superior performance in terms of accuracy and equitable tutoring compared to larger models like ChatGPT, despite the latter being pre-trained on a significantly larger dataset with a higher parameter count. These findings suggest that while GPT-series models generally exhibit strong performance, smaller fine-tuned models or even traditional machine learning (ML) methods can sometimes achieve better results on certain tasks. Therefore, combining these approaches—such as fine-tuning GPT models or integrating outputs from traditional ML methods—may yield improved performance over relying on a single method.

Lastly, two research exclusively utilized other LLMs. [50] fine-tuned the RoBERTa-Longformer model to extend the input context window for classification tasks aimed at providing formative feedback in intelligent textbooks. [3] fine-tuned the LLaMA2-7B model, expanding a set of physics problems from an initial count of 766 to 7,983, thereby enriching educational resources.

## 4.3 Implementation of LLM-based systems in K-12 STEM education (RQ 1.3)

In our study, we identified 9 articles that specifically investigated interactions between LLMs and human participants. [73] developed a web interface enabling teachers on Prolific to interact with AI-simulated students, effectively generating teacher-student interaction data. [69] implemented a web platform for Amazon Mechanical Turk participants to engage with Socratic sub-questions generated by GPT-3, which improved the success rates of their study. [7] utilized ChatGPT to facilitate student discussions on quantum physics, resulting in improved student perceptions and a more cautious approach to using ChatGPT after the intervention. [82] conducted a quasi-experiment where GPT-3 was integrated with iPads via LINE, significantly enhancing students' intrinsic motivation and self-regulation in mathematics learning. Five additional studies developed demonstration interfaces but did not conduct user studies or evaluate effectiveness. For example, [84] and [54] integrated GPT-4 for personalized tutoring. Meanwhile, [2, 16, 53] developed demo interfaces based on their research findings.

## 4.4 Evaluation methods to assess the performance of prompting strategies (RQ 2.1)

There are 8 papers [15, 21, 23, 26, 38, 57, 71, 84] that offer no comparison; they either present a prompt or an application and simply demonstrate its potential. Additionally, 13 studies compare a single prompt with either a dataset or human performance to assess the extent to which large language models (LLMs) achieve high performance [18, 42–44, 53, 54], as well as which LLM performs better in various contexts [27, 42, 50, 52, 70, 73, 76]. Only 9 articles [16, 27, 42, 50, 52, 69, 70, 73, 76] establish multiple prompting strategies and conduct comparisons to evaluate which prompting methods yield better results.

*4.4.1 Evaluation of prompt engineering strategies.* A total of 18 articles evaluate the LLM's performance across various prompting strategies [2, 7, 15, 16, 18, 26–28, 36, 38, 42, 44, 57, 70, 71, 73, 76, 84]. In addition, because of the non-determinism of LLM outputs, repeated independent interactions were adopted in 17 articles to obtain a more reliable assessment of the outputs [2, 13, 18, 21, 27, 36, 42–44, 50, 52–54, 57, 62, 73, 76]. For example, [2] developed an interface allowing users to interact with the chatbot and collected data for independent evaluation, with each prompt running three outputs to compare the effectiveness of different custom-designed prompt personalities. Similarly, [21] had ChatGPT solve the same problem 60 times and found that it consistently answered only three questions correctly every time, with a degree of randomness in accuracy for the remaining cases.

*4.4.2 Temperature.* Some articles mention the issue of temperature, with only two studies [13, 53] explicitly stating that they set the temperature to zero during evaluation. [13] found that a higher temperature for GPT-3.5 for data augmentation on stimulated students yields comparable results to a lower temperature. The benefit was generating more diverse student responses, reducing overfitting even with more augmentation data points. In fact, according to OpenAI's documentation[1], even when the temperature is set to 0, there will still be some variability in the results, though with a lower probability. Therefore, thoroughly evaluating multiple times may be more effective than simply lowering the temperature.

*4.4.3 Reproducibility.* Regarding the reproducibility of the findings in reviewed articles, we observed that 12 articles provided prompts that are accessible, replicable, and reproducible [16, 18, 27, 28, 42–44, 54, 57, 62, 76, 84]. Then, 11 articles partially shared their prompts [2, 15, 21, 23, 26, 36, 38, 53, 70, 73, 82]. 2 articles [3, 13] provided prompt templates, offering an outline of the prompt's structure. Additionally, 1 article included the prompt in an image format, which may limit its accessibility for replication purposes [7]. Another study did not make its prompt available [69], and in one case, the prompt was intended to be accessible via GitHub, but the link was unfortunately inactive at the time of review [52]. By reviewing 30 articles in K-12 STEM education, roughly 40% (12 out of 30 articles) demonstrated prompts that are accessible and replicable, while the others lack detailed prompts and transparent methodologies, which hinder the reproducibility of studies. We acknowledge that there may be various reasons behind the limited availability of prompts. However, without access to the exact prompts used, it is challenging for educational practitioners to effectively apply LLMs to the educational task and other researchers to replicate the results or build upon the work. This issue is compounded by the non-deterministic nature of LLM outputs, as minor variations in prompts or model parameters can lead to different outcomes. Thus, we encourage future studies to consider sharing their materials in a fully accessible manner to promote greater reproducibility in this important area of research.

## 4.5 Metrics to measure the performance of prompting strategies with LLMs (RQ 2.2)

Our findings highlight the diverse evaluation methods employed by various researchers. 14 articles utilize custom indicators [2, 15, 16, 18, 21, 23, 36, 38, 42, 54, 57, 62, 69, 73] such as determining whether a given lesson plan aligns with the 5Es instructional model [26]. Some of these custom indicators are evaluated post-hoc by human coders, who calculate consensus scores. The resulting Cohen's κ values range between 0.72 and 0.89, as demonstrated in [2, 21, 46, 62]. In other cases, machine learning models or predefined rules are used to compute the indicators, followed by a direct comparison of numerical values. For instance, [54] employs a BERT model along with several readability index models in regression analysis to assess the readability of the generated outputs. Moreover, there are 15 articles that evaluate the prompting strategy and LLM performance based on traditional machine learning metrics, such as accuracy, F1 Score, and precision [3, 13, 18, 21, 28, 36, 43, 44, 50, 52, 53, 57, 62, 73, 76]. Then, some studies incorporate statistical metrics such as mean and standard deviation in their evaluations to analyze the distribution of GPT's performance [24, 66]. For example, [21] found that although GPT's accuracy in solving physics problems was comparable to that of students, the actual performance distribution was much narrower than that of students. Moreover, GPT exhibited particularly strong performance on some questions while performing poorly on others, in contrast to students, whose errors were more evenly distributed across different problems.

---
[1]https://platform.openai.com/docs/api-reference

### 4.6 Effectiveness of LLM-based system in real-world educational settings (RQ 2.3)

Only 2 articles have conducted real-world experiments with students to validate LLM-based prompting strategies in educational settings [7, 82]. The study by [7] involved a pilot study in a secondary school physics classroom, where 53 students participated in a two-lesson intervention using ChatGPT. The study found that the integration of ChatGPT had a positive impact on students' perceptions of the tool, with increased agreement on the benefits of using ChatGPT in daily life and education. Students also appreciated its potential for supporting critical thinking, especially in abstract subjects like quantum physics. The study by [82] focused on using a ChatGPT-based Intelligent Learning Aid (CILA) system in a blended learning environment to promote self-regulation and knowledge construction in a mathematics class. This study involved 70 students and compared the effectiveness of CILA system with traditional search engines like Google. Findings indicated that CILA significantly enhanced students' self-regulation and knowledge construction, particularly by providing real-time, convergent information that minimized learning interruptions. Other papers have not directly engaged with real users; instead, they rely on expert reviews, individual researcher judgments, theoretical frameworks, or machine learning models as the basis for their findings.

### 4.7 Limitations of using LLMs with prompting strategies in previous studies (RQ 3)

Despite the promising advancements in leveraging LLMs with prompting strategies for K-12 STEM education, the research we reviewed indicated several limitations related to prompt engineering.

*4.7.1 Complexity in Designing Effective Prompts.* Designing prompts that effectively elicit the desired responses from LLMs is a complex task that often requires iterative refinement and domain expertise. [2] emphasized the need for clear and precise prompts to maximize the benefits of using LLMs as teaching assistants. However, the process of crafting such prompts can be time-consuming and may require specialized knowledge in both the subject matter and prompt engineering techniques. This complexity can be a barrier for educators who may not have the resources or expertise to develop and evaluate optimal prompts for their specific educational needs.

*4.7.2 Sensitivity to Prompt Variations.* One key limitation is the sensitivity of LLMs to minor changes in prompts, which can lead to very different outputs and affect the reliability of results. For instance, [21] found that ChatGPT's performance on physics problems was highly sensitive to subtle modifications in the prompt wording, resulting in varying levels of accuracy. This sensitivity complicates the design of effective prompts, as even minor adjustments can significantly impact the model's responses, making it challenging to achieve consistent performance across different tasks or educational contexts.

*4.7.3 Lack of Prompt Generalizability and Scalability.* The effectiveness of specific prompting strategies may not be generalized across different subjects or tasks. As observed by [70], prompting techniques that worked well for text-based programming did not yield similar success in visual programming tasks. Similarly, [28] noted that simple few-shot prompts with limited examples were insufficient for capturing nuanced criteria in evaluating the tutor's praise, suggesting that more elaborate and domain-specific prompts are necessary. [36] also noted that CoT without domain-specific information leads to limited improvements, as LLMs struggle to break down tasks effectively. This limitation indicates that prompt engineering often requires significant customization for each specific application, limiting the scalability of these strategies.

*4.7.4 Dependence on Model-Specific Features.* Prompting strategies sometimes rely on features specific to certain LLMs, limiting their applicability to other models. For example, some advanced prompting techniques may leverage the large context window of GPT-4, which is not available in smaller or open-source models [27, 52]. This dependence on proprietary models raises concerns about accessibility and equity, as not all educators or researchers have the resources to access the latest or most capable LLMs. Additionally, fine-tuned smaller models sometimes outperform larger models when properly fine-tuned and prompted [73], suggesting that prompt engineering must be tailored not only to the task but also to the specific model being used.

*4.7.5 Limitations in Handling Visual Educational Content.* Prompting strategies may not sufficiently overcome LLM limitations in handling visual educational content. For instance, [21] observed that despite using chain-of-thought prompting, LLMs struggled with interpreting kinematic graphs in physics problems. Similarly, [84] found that GPT-4 effectively solves algebra problems but struggles with geometric questions involving visual elements like graphs and diagrams, highlighting limitations in visual comprehension. These findings suggest that while prompting strategies can enhance LLM performance, they may not be sufficient to address fundamental limitations in the models' ability to handle certain educational content.

*4.7.6 Non-determinism and Reproducibility Issues.* The inherent non-determinism of LLM outputs poses challenges for reproducibility. As highlighted by [54], LLM's responses can vary due to its random nature, making it difficult to replicate results consistently. This stochastic nature requires multiple runs to obtain reliable assessments, as single outputs may not be representative of the model's typical performance [70]. Moreover, the lack of detailed documentation of prompts in some studies further hinders reproducibility, as other researchers cannot accurately replicate the prompting strategies used.

*4.7.7 Evaluation Challenges Specific to Prompting Strategies.* The methods used to evaluate the effectiveness of LLMs with prompting strategies in education often have limitations. Many studies rely on subjective assessments or custom indicators without standardized benchmarks, which hinders comparability across studies. For example, [26] used qualitative analysis and subjective evaluation of ChatGPT-generated lesson plans, which may vary greatly due to diverse teaching standards. Similarly, [13, 57] noted the need for publicly available benchmark datasets to compare and evaluate AI-generated outputs with different prompting strategies systematically.

## 5 DISCUSSION AND CONCLUSION
### 5.1 Summary of Key Findings
Our study investigated the application of prompt engineering strategies with Large Language Models (LLMs) in K-12 STEM education. Our analysis focused on three primary research questions: first, identifying the prompting strategies employed in various educational tasks across different educational levels within K-12 settings; second, exploring the types of LLMs utilized as the backend model for these prompting strategies; and third, evaluating the methods researchers have implemented to assess the effectiveness of LLM-facilitated applications. Through this investigation, we aim to shed light on how prompt engineering can enhance educational outcomes and offer insights into the practical use of LLMs in real-world K-12 STEM education.

*Firstly*, we identified a variety of prompting strategies utilized in the literature, including simple prompting (or direct instruction prompting), zero-shot prompting, few-shot prompting, and chain-of-thought prompting. Simple prompting and zero-shot prompting were the most commonly used strategies due to their simplicity and ease of implementation. However, more sophisticated strategies like few-shot prompting and chain-of-thought were found to enhance the performance of LLMs by providing additional context or examples, thereby improving their ability to handle complex educational tasks (e.g., Problem Solving [2, 3, 15, 18, 21, 23, 27, 38, 42–44, 52, 53, 62, 70, 82] and Analyzing and Labeling[28, 42, 53, 76]. For more, please refer to Table 2). *Secondly*, the majority of studies employed models from the GPT series, such as GPT-3, GPT-3.5, and GPT-4, which demonstrated strong performance across various educational tasks. However, other studies have also explored the use of different LLMs beyond the GPT series, including fine-tuned versions of LLaMA and open-source models like Blender, BLOOM, and YouChat. In certain cases, smaller, fine-tuned models outperformed larger pre-trained ones, suggesting that model selection should be tailored to the specific educational task and context. *Lastly*, researchers used a range of evaluation methods to assess the effectiveness of prompting strategies, including comparisons with human performance, the use of custom indicators and metrics, machine learning evaluation metrics (e.g., accuracy, F1 score), and statistical analyses of LLM outputs. However, only a limited number of studies conducted real-world experiments involving interactions with actual students or educators, revealing a gap in empirical validation and practical implementation that future research should aim to address.

### 5.2 Discussion

The findings of our study underscore the critical role of prompt engineering in enhancing the effectiveness of LLMs in K-12 STEM education. Advanced prompting strategies such as few-shot prompting and chain-of-thought prompting provide LLMs with additional guidance, enabling them to better understand complex tasks and generate more accurate and coherent responses.

The predominance of GPT-series models in the reviewed studies reflects their accessibility and robust performance across a range of tasks. However, the success of smaller, fine-tuned models in certain contexts suggests that customization and fine-tuning can be effective strategies, particularly when resources are limited or specific domain expertise is required. This highlights the importance of selecting LLMs to align with the educational objectives and constraints of specific settings.

The evaluation methods employed in the studies varied widely, with many relying on custom indicators or comparisons with existing datasets. The lack of standardized benchmarks and limited use of empirical validation present challenges for assessing the true effectiveness of LLMs in educational contexts. The few studies that involved direct interactions with students or educators provided valuable insights into the practical implications of using LLMs in classrooms but also pointed to the need for more rigorous and comprehensive evaluations.

Several limitations were identified in the current research. Many studies focused on narrow domains or specific datasets, limiting the generalizability of their findings. Methodological constraints, such as small sample sizes, lack of control groups, and absence of pre-tests, further hinder the robustness of conclusions. Additionally, inherent limitations of LLMs, such as variability in outputs and difficulty handling visual information, pose challenges for their integration into STEM education. The sensitivity of LLMs to prompt changes and the non-deterministic nature of their outputs necessitate careful design and testing of prompts to ensure consistency and reliability.

## 5.3 Recommendations for Future Works

**Empirical Validation in Real-world Settings:** There is a pressing need for more empirical studies that implement LLMs with advanced prompting strategies in actual classroom environments. Such studies should involve different prompts, diverse educational contexts, subjects, and student populations to assess the best practice, generalizability, and practical impact of LLM with prompt engineering on learning outcomes.

**Development of Dataset and Evaluation Metrics for Prompt Engineering:** Creating standardized datasets, benchmarks, and metrics for evaluating LLM performance with various prompts in educational tasks would ensure consistent, comparable results. This could involve building datasets across STEM subjects and educational levels, allowing researchers to rigorously assess different prompting strategies and models.

**Customization and Fine-tuning of LLMs with augmented data by LLM with prompt engineering:** Exploring the customization and fine-tuning of LLMs for specific educational purposes could enhance their effectiveness and accessibility. Investigating the potential of smaller, open-source models that can be adapted to meet the needs of different educational environments is a promising avenue for making AI tools more widely available and relevant to educators and students.

**Addressing Methodological and Ethical Challenges with Prompt Engineering:** Future research should aim to overcome the methodological limitations identified, such as enhancing study designs with larger sample sizes, control groups, and pre- and post-assessments. Additionally, ethical considerations, including issues related to bias, data privacy, and the evolving role of educators in the context of AI integration, should be thoroughly examined to ensure responsible and equitable use of LLMs in education.

**Enhancing Multimodal Capabilities of LLMs with Prompt Engineering:** Given the importance of visual information in STEM education, improving the ability of LLMs to process and generate content that includes graphs, diagrams, and other visual elements is essential. Research into multimodal AI models that can handle both textual and visual data could significantly expand the applicability of LLMs in STEM subjects.

**Longitudinal Studies on Learning Outcomes with LLMs with different prompts:** Conducting longitudinal studies to evaluate the sustained impact of LLMs with different prompts on student learning, engagement, and motivation would provide deeper insights into the long-term benefits and potential drawbacks of integrating AI technologies into educational practices.

By addressing these areas, future work can contribute to the use of LLMs and Prompt Engineering in the development of effective, ethical, and accessible AI-driven tools that enhance STEM education and support educators in fostering deeper understanding and engagement among students.